\documentclass[11pt]{article}
\usepackage[margin=1in]{geometry}
\usepackage[T1]{fontenc}
\usepackage{lmodern}
\usepackage{microtype}
\usepackage{amsmath,amssymb}
\usepackage{booktabs}
\usepackage{multirow}
\usepackage{graphicx}
\usepackage{xcolor}
\usepackage{caption}
\usepackage{tikz}
\usetikzlibrary{positioning,arrows.meta,fit,calc}
\usepackage[numbers,sort&compress]{natbib}
\usepackage{xurl}
\usepackage[colorlinks=true,linkcolor=blue!50!black,citecolor=blue!50!black,urlcolor=blue!50!black]{hyperref}
\usepackage{enumitem}
\usepackage[section]{placeins}%
\setlist{nosep,leftmargin=*}
\newcommand{\bevl}{Beveridge-Large-C}
\newcommand{\bevp}{Beveridge-Pro-C}
\newcommand{\bevw}{Beveridge-Pro-C-Web}
\newcommand{\best}[1]{\textbf{#1}}

\title{\textbf{Beveridge: Open Weight System One Decision Models\\with Calibrated Probabilities}}
\author{Sathiskumar Jothi\\[2pt]
\small BDST: Big Data Science \& Technology Ltd, United Kingdom\\
\small \href{mailto:s.jothi@bdst.uk}{\texttt{s.jothi@bdst.uk}}\quad \url{https://bev.sjothi.com}}
\date{October 2026}

\begin{document}
\maketitle

\begin{abstract}
Many production systems need a model to make a narrow decision about a document (route a ticket, check a limit,
refuse a request, pick the next step of a browser agent) rather than to write text about it. We present
\emph{Beveridge}, a family of open-weight \emph{System One} decision models that take a document and up to 15
typed questions (yes/no, choice among described options, or a level on an ordered scale) and return, in one forward
pass without generating text, an answer with a calibrated probability for every option. \bevl{} (398M parameters)
adds a 3.5M-parameter decision network to ModernBERT-large: typed number and date features in the input, exact facts
from a rule engine, question-to-document evidence attention, a three-step recursive core with learned halting and an
exact yes/no path. \bevp{} adapts Gemma~4 26B-A4B with LoRA to score option letters, and \bevw{} extends it to the next
step of a browser agent. On held-out tests the 0.4B model reaches 93.0\% on our operations-document benchmark against
91.1\% for the best of five general LLMs (31B--675B) and 87.9\% for a commercial System One API, and, because its
confidences are calibrated (ECE 0.029 on that benchmark), it can hand over 76\% of decisions automatically at 95\%
accuracy, where no general LLM we tested can hand over any. It answers in 4.9\,ms per decision on one NVIDIA H200 GPU and
60\,ms on the 14 core CPU of an Apple M4 Pro laptop. \bevw{} raises next-step accuracy on web pages from unseen site types from 49.4\% to 98.2\% and completes
live browsing tasks without losing document accuracy. We describe the models, the training recipes and an evaluation
on 19 held-out test sets against a commercial System One API and
five general LLMs, including where the models are weak. Weights
and inference code are released under Apache-2.0.
\end{abstract}

\begin{center}\small
\begin{tabular}{@{}ll@{}}
\toprule
\multicolumn{2}{@{}l}{\textbf{Models, code and videos}}\\
\midrule
\bevl{} & \href{https://huggingface.co/sathiskumarjothi0/Beveridge-Large-C}{\texttt{huggingface.co/sathiskumarjothi0/Beveridge-Large-C}}\\
\bevp{} & \href{https://huggingface.co/sathiskumarjothi0/Beveridge-Pro-C}{\texttt{huggingface.co/sathiskumarjothi0/Beveridge-Pro-C}}\\
\bevw{} & \href{https://huggingface.co/sathiskumarjothi0/Beveridge-Pro-C-Web}{\texttt{huggingface.co/sathiskumarjothi0/Beveridge-Pro-C-Web}}\\
Python package (PyPI) & \href{https://pypi.org/project/bev-decisions/}{\texttt{pypi.org/project/bev-decisions}} (\texttt{pip install bev-decisions})\\
Same package, short name & \href{https://pypi.org/project/beveridge/}{\texttt{pypi.org/project/beveridge}} (\texttt{pip install beveridge})\\
Code and examples (GitHub) & \href{https://github.com/sathiskumarjothi0-oss/bev-decisions-pypi}{\texttt{github.com/sathiskumarjothi0-oss/bev-decisions-pypi}}\\
Release used in this paper & \href{https://github.com/sathiskumarjothi0-oss/bev-decisions-pypi/releases/tag/v0.2.5}{\texttt{v0.2.5}} (GitHub release and PyPI)\\
Website & \href{https://bev.sjothi.com/research/}{\texttt{bev.sjothi.com/research}}\\
Videos & \href{https://bev.sjothi.com/research/media/beveridge-web-browsing.mp4}{web browsing},
\href{https://bev.sjothi.com/research/media/beveridge-voice-examples.mp4}{voice assistant},
\href{https://bev.sjothi.com/research/media/bev-snake.mp4}{game agent} (Figure~\ref{fig:demos})\\
\bottomrule
\end{tabular}
\end{center}

\section{Introduction}

Large language models (LLMs) are now routinely used as classifiers: a prompt describes a document and a question, and
the generated text is parsed into a label. This works, but it is an obstinate fit for automation. The answer has to be
parsed from free text, the model's stated confidence is not tied to how often it is right, one decision costs seconds
and a remote call; and the same weights that make the model fluent make it slow. Following the distinction between
fast, automatic judgement and slow, deliberate reasoning \citep{kahneman2011thinking}, we call a model that answers
typed questions in one pass, with a probability for every option and no text generation, a \emph{System One} decision
model.

\emph{Beveridge} is a family of such models with open weights. A request is a document (any text: an e-mail, a log, an
invoice, a web page as JSON) and named, typed questions; the response is a probability distribution per question
(Section~\ref{sec:task}). Three models are released:
\begin{itemize}
  \item \textbf{\bevl{}} (0.4B): ModernBERT-large \citep{warner2024modernbert} with a decision network that reads typed
  values and exact facts, refines its answers over three recursive steps and answers all questions of a request in one
  encoder pass. It runs on a laptop.
  \item \textbf{\bevp{}} (25.8B total, about 4B active per token): Gemma~4 26B-A4B \citep{gemma4} with a LoRA
  adapter \citep{hu2021lora} trained to score option letters, for knowledge heavy decisions on one GPU.
  \item \textbf{\bevw{}}: \bevp{} with a second round of adapter training on generated web navigation steps, used as the
  decision maker of a browser agent.
\end{itemize}

\paragraph{Contributions.} The typed question interface follows an existing System One API \citep{typesafejev}, and most
components we use are known; our contributions are the following.
\begin{enumerate}
  \item An encoder decision architecture that combines typed value features, an exact fact engine and a halting recursive
  core, with a closed form yes/no path (Eq.~\ref{eq:yesno}) that lets an exact comparison decide when it applies and
  inference time ablations showing how much the trained model relies on each part (Section~\ref{sec:ablation}).
  \item Open-weight, calibrated decision models (Apache 2.0) and their training recipes (Section~\ref{sec:training}).
  \item A comparison on 19 held out test sets against a commercial System One model and five general LLMs (31B--675B)
  under identical inputs, showing that a 0.4B calibrated encoder is the most accurate system on operations document
  decisions and that, measured as automation at a target accuracy, none of the general LLMs can hand over any decision at
  95\% accuracy because of how they report confidence (Sections~\ref{sec:eval}--\ref{sec:results}).
  \item The use of a System One model as the decision maker of a browser agent (Section~\ref{sec:web}).
\end{enumerate}

\section{Related work}\label{sec:related}

\paragraph{LLMs as classifiers and their confidence.} Prompted LLMs classify text zero or few shot
\citep{brown2020gpt3}, but their label probabilities are biased by the prompt format \citep{zhao2021calibrate}, and
confidences they state in words are often overconfident \citep{xiong2024uncertainty,tian2023justask}, even though token
probabilities of base models can be well calibrated \citep{kadavath2022know}. Our LLM baselines state probabilities in
JSON, as an automation pipeline would use them; \bevp{} instead reads the probabilities of option letters at one
position, and both Beveridge models are calibrated after training.

\paragraph{Calibration and selective prediction.} Temperature scaling \citep{guo2017calibration} and the expected
calibration error \citep{naeini2015ece} are standard, and fine-tuned transformer encoders are known to be relatively
well calibrated \citep{desai2020calibration}. Handing over only confident decisions is selective classification
\citep{elyaniv2010selective,geifman2017selective}; our \emph{automation at a target accuracy} is one point of its
risk coverage curve, with tied confidences handed over together.

\paragraph{Numbers in language models.} Subword embeddings encode numbers poorly \citep{wallace2019numbers,
thawani2021numbers}; continuous number encodings such as xVal \citep{golkar2023xval} add the value to the representation.
Our value features follow this line and keep the digits as text as well, for money, dates, times and plain numbers.

\paragraph{Neural models with exact operations.} Table parsers aggregate selected cells with exact operators
\citep{herzig2020tapas}, NumNet reasons over number graphs \citep{ran2019numnet}, neural arithmetic units learn exact
operations \citep{trask2018nalu}, and LLMs call tools \citep{schick2023toolformer}. Beveridge computes a fixed set of
facts with a rule engine and lets the network select among them, including the closed form yes/no mixture of
Eq.~\ref{eq:yesno}.

\paragraph{Adaptive computation.} Adaptive computation time \citep{graves2016act}, Universal Transformers
\citep{dehghani2019universal} and PonderNet \citep{banino2021pondernet} apply shared layers repeatedly with learned
halting, our recursive core is a three step instance over question and fact vectors.

\paragraph{Encoders, routing and agents.} Bidirectional encoders \citep{devlin2019bert,warner2024modernbert} remain the
efficient choice for classification, and zero shot classification can be framed as entailment \citep{yin2019zeroshot};
Beveridge instead answers several typed questions with written option criteria in one pass. Cascades and routers send
requests to cheaper or stronger models \citep{chen2023frugalgpt,ong2024routellm}; our router uses calibrated confidence.
Guardrail classifiers such as Llama Guard \citep{inan2023llamaguard} decide request safety; we include this as one
decision type. Web agents are evaluated on WebShop, Mind2Web and WebArena \citep{yao2022webshop,deng2023mind2web,
zhou2024webarena}; we evaluate a single decision model inside an existing browser loop \citep{jevultrafast}, on generated
held out sites and two live tasks only.

\section{Typed decisions}\label{sec:task}

A \emph{request} is a pair (\emph{state}, \emph{questions}). The state is a document. Each of up to 15 questions has a name, an instruction and a type. In \bevl{} the questions, their options and the
state share one input of 4,096 tokens (the state receives the tokens the questions leave, Section~\ref{sec:large}), in
\bevp{} each question gets its own 4,096 token input (Section~\ref{sec:pro}). Training and
evaluation used one to five questions per request:
\begin{itemize}
  \item \texttt{noul} (yes/no): the answer is $P(\text{yes})$.
  \item \texttt{choice}: one of $K$ options, each given as a key and a written criterion (up to 300 options), the answer
  is a distribution over the keys.
  \item \texttt{score}: a level on an ordered scale whose levels carry written criteria, lowest first; the answer is a
  distribution over levels and its expected level.
\end{itemize}
A response gives, per question, the most likely option, its probability (the \emph{confidence}) and the full
distribution. Appendix~\ref{app:example} shows a request. We require three properties: (i) no text is generated, so
nothing needs parsing and the answer is always one of the offered options; (ii) confidences are calibrated, so a
threshold on confidence controls the error rate of automated decisions; (iii) answers do not depend on the order in
which questions or options are listed.

The format matches the System One interface of TypeSafe's Jev \citep{typesafejev}, which lets us evaluate both on
identical requests.

\section{Models}\label{sec:models}

\subsection{\bevl{}}\label{sec:large}

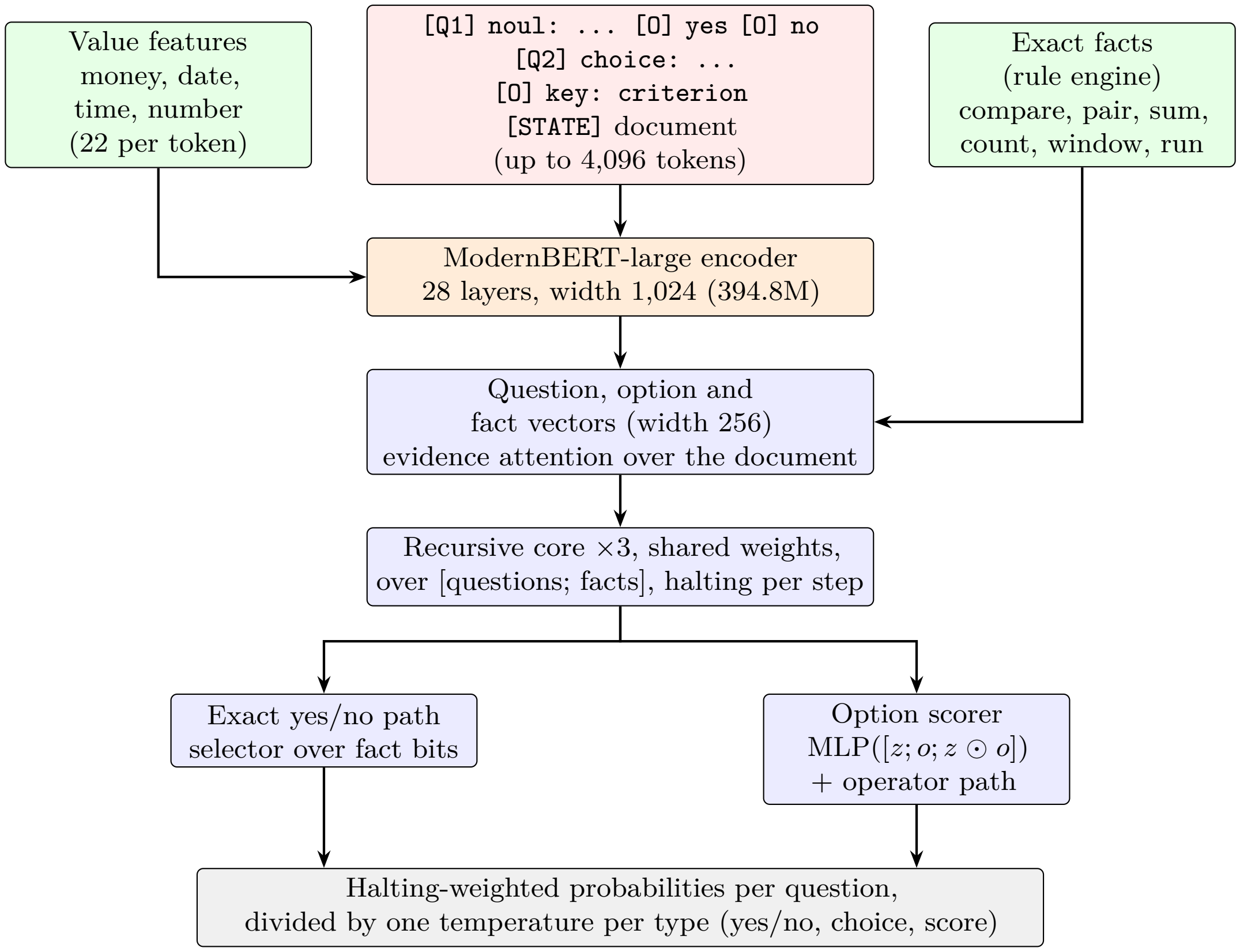
\begin{figure}[htbp]
\centering
\begin{tikzpicture}[
  box/.style={draw, rounded corners=2pt, align=center, font=\footnotesize, minimum height=8mm, inner sep=3pt},
  inp/.style={box, fill=red!8}, enc/.style={box, fill=orange!15}, dec/.style={box, fill=blue!8},
  ex/.style={box, fill=green!10}, outp/.style={box, fill=gray!12},
  ar/.style={-{Stealth[length=2mm]}, thick}, node distance=6mm]
\node[inp, text width=5.6cm] (txt) at (0,0) {\texttt{[Q1] noul: ... [O] yes [O] no}\\\texttt{[Q2] choice: ... [O] key: criterion}\\\texttt{[STATE]} document (up to 4,096 tokens)};
\node[ex, text width=3.3cm] (val) at (-5.3,0) {Value features\\money, date, time, number\\(22 per token)};
\node[ex, text width=3.3cm] (fact) at (5.3,0) {Exact facts (rule engine)\\compare, pair, sum,\\count, window, run};
\node[enc, text width=5.6cm, below=of txt] (encd) {ModernBERT-large encoder\\28 layers, width 1,024 (394.8M)};
\node[dec, text width=5.6cm, below=of encd] (vec) {Question, option and fact vectors (width 256)\\evidence attention over the document};
\node[dec, text width=5.6cm, below=of vec] (core) {Recursive core $\times$3, shared weights,\\over [questions; facts], halting per step};
\node[dec, text width=3.3cm, anchor=north] (yn) at ($(core.south)+(-3.4,-1.0)$) {Exact yes/no path\\selector over fact bits};
\node[dec, text width=3.3cm, anchor=north] (opt) at ($(core.south)+(3.4,-1.0)$) {Option scorer\\$\mathrm{MLP}([z;o;z\odot o])$\\+ operator path};
\node[outp, text width=9.5cm, anchor=north] (o) at ($(core.south)+(0,-3.0)$) {Halting-weighted probabilities per question,\\divided by one temperature per type (yes/no, choice, score)};
\draw[ar] (txt) -- (encd);
\draw[ar] (val.south) |- (encd.west);
\draw[ar] (fact.south) |- (vec.east);
\draw[ar] (encd) -- (vec);
\draw[ar] (vec) -- (core);
\draw[ar] (core.south) -- ++(0,-4mm) -| (yn.north);
\draw[ar] (core.south) -- ++(0,-4mm) -| (opt.north);
\draw[ar] (yn.south) -- (yn.south |- o.north);
\draw[ar] (opt.south) -- (opt.south |- o.north);
\end{tikzpicture}
\caption{\bevl{}. All questions and their options are placed before the document and read in one encoder pass; a
rule engine supplies typed value features to the encoder input and exact facts to the decision network.}
\label{fig:large}
\end{figure}

\paragraph{Input.} All questions and their options are written before the document as one sequence:
\texttt{[Q$i$] type: instruction [O] key: criterion \ldots{} [STATE] document}. Yes/no questions receive two options,
\emph{yes} and \emph{no}, so that all types share one scoring mechanism. Putting questions first guarantees they are
never truncated, the document receives the remaining tokens and is cut at the 4,096 token limit. If the questions would
leave fewer than 160 tokens for the document, the longest option texts are shortened first, and option sets that still do
not fit (very large intent lists) are encoded separately and attached late (\emph{late interaction}). Document tokens
$= 4{,}096 -$ question and option tokens, for example, five questions of about 60 tokens each leave
$4{,}096 - 300 = 3{,}796$ tokens for the document. The number of questions is bounded separately by the learned
question index embeddings (16 slots, one of which tags facts shared by all questions), so a request holds at most 15
questions however short they are.

\paragraph{Typed values.} A deterministic parser finds money amounts, dates, times and plain numbers in the state, the
instructions and the options. Each value becomes 22 features (a one hot kind, a signed log magnitude, a date ordinal and
16 Fourier features of the log magnitude), projected to the encoder width by a zero-initialised linear layer and added to
the embeddings of the value's tokens. The model therefore sees ``\$9,412.50'' both as text and as the number 9412.5.

\paragraph{Exact facts.} The same parser feeds a rule engine that computes facts relevant to each question:
comparisons between two values (\texttt{cmp}, \texttt{pair}), sums of line items (\texttt{sum}), counts of items
satisfying a condition (\texttt{count}), values within a time window (\texttt{window}) and runs of events
(\texttt{run}). Each fact has eight numeric features, its kind, an exact yes/no bit where it applies and, for counts,
the exact result. The engine has no learned parameters.

\paragraph{Encoder.} ModernBERT-large: 28 layers, hidden width 1,024, 16 heads, alternating global attention every third
layer with 128 token local windows elsewhere, rotary positions \citep{warner2024modernbert}. All 398.3M parameters are
fine-tuned.

\paragraph{Decision network (3.5M parameters).} Each question vector is the mean of its tokens projected to width 256,
plus learned question index and type embeddings, it then reads the whole document once more through evidence attention.
Option vectors are the projected means of each option's tokens. Fact vectors combine the fact features with the encoder
states of the two text spans the fact refers to. A single transformer layer (8 heads, width 256) is applied three times
with shared weights over the concatenation of question and fact vectors, after each step $k$ a gated update gives the
question state $z_k$ and a halting head gives the probability of stopping at that step, in the manner of adaptive
computation \citep{graves2016act,banino2021pondernet}. A direct path also lets each question attend to its own facts.

For \texttt{choice} and \texttt{score} questions the logit of option $o$ at step $k$ is
$\mathrm{MLP}([z_k; o; z_k \odot o])$ plus a gated operator term built from lexical option features and the match
between an option's number and a computed fact result (for counting questions). For \texttt{noul} questions the base
logit is the difference between the scores of the \emph{yes} and \emph{no} options, and an exact path mixes in the facts:
\begin{equation}
P(\text{yes}) \;=\; a_0\,\sigma(b) \;+\; \sum_{j} a_j\,\sigma\!\big(b + w\,(2\,\beta_j - 1)\big),
\label{eq:yesno}
\end{equation}
where $b$ is the base logit, $\beta_j \in \{0,1\}$ the exact bit of fact $j$, $a$ a learned selector distribution over
``no fact'' and the question's facts, and $w$ a learned signed strength (initialised at 2). When a fact answers the
question exactly (``Is the total over the limit?''), the selector can put its weight on it; otherwise $a_0$ keeps the
learned answer. Example: $b=0.4$, $a_0=0.2$, $a_1=0.8$ on a fact with $\beta_1=1$, $w=2$ gives
$0.2\,\sigma(0.4)+0.8\,\sigma(2.4)=0.2\cdot0.599+0.8\cdot0.917=0.853$.

\begin{figure}[htbp]
\centering
\includegraphics[width=\linewidth]{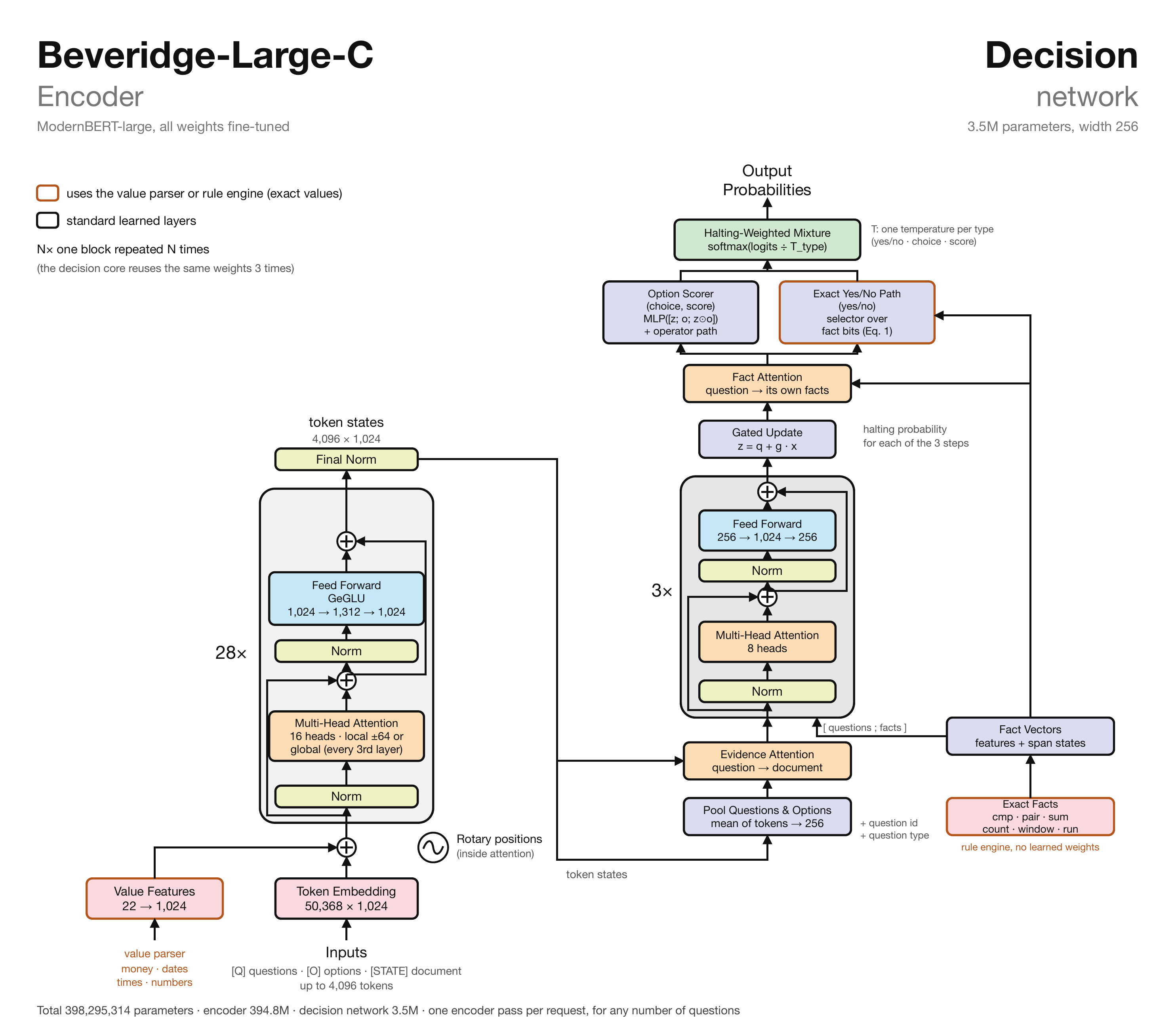}
\caption{\bevl{} layer by layer. Left, the ModernBERT-large encoder (28 pre-norm layers), with typed value features
added to the token embeddings. Right, the decision network. Questions and options are pooled from the token states and
read the document again through evidence attention, one transformer layer (width 256) is applied three times with shared
weights over questions and facts, each step giving a gated question state and a halting probability, and each question then
attends to its own facts and is scored by the option scorer or the exact yes/no path (Eq.~\ref{eq:yesno}). Orange
outlines mark the parts that use the value parser or the rule engine.}
\label{fig:block}
\end{figure}

The final distribution of a question is the halting weighted mixture of its per step distributions, each computed from
logits divided by a temperature fitted per question type after training (Section~\ref{sec:calib}). Table~\ref{tab:params}
lists the parameter budget and Figure~\ref{fig:block} shows every layer. A request with any number of questions costs
one encoder pass.

\begin{table}[htbp]
\centering\small
\resizebox{\linewidth}{!}{%
\begin{tabular}{lrl}
\toprule
Component & Parameters & Role \\
\midrule
ModernBERT-large encoder & 394,781,696 & reads questions, options and document together \\
Value projection & 23,552 & 22 value features $\to$ 1,024, added to token embeddings \\
Question vectors, ids, types & 267,264 & question representation \\
Evidence attention & 656,128 & each question reads the document again \\
Option vectors (in context, late) & 524,800 & option representation \\
Fact vectors & 593,920 & facts + encoder states of their spans \\
Recursive core, gate, halting & 790,018 & three shared steps over questions and facts \\
Direct fact path & 196,864 & question-to-own-facts attention \\
Option scorer, operator path & 197,385 & option logits \\
Exact yes/no path & 197,635 & selector over fact bits, Eq.~\ref{eq:yesno} \\
Other (legacy yes/no head, temperatures) & 66,052 & \\
\midrule
Total & 398,295,314 & \\
\bottomrule
\end{tabular}}
\caption{\bevl{} parameters, measured on the instantiated model.}
\label{tab:params}
\end{table}

\subsection{\bevp{}}\label{sec:pro}

\bevp{} uses an instruction tuned mixture of experts LLM, Gemma~4 26B-A4B \citep{gemma4}, as a scorer rather than a
generator. Each question becomes one prompt of up to 4,096 tokens in the model's chat format, so the document is read
once per question:
\begin{quote}\small
\begin{verbatim}
Document:
<state, cut to fit>

Question: <instruction>
Options:
A. <option>
B. <option> ...
Answer with the letter only.
\end{verbatim}
\end{quote}
Yes/no questions get the options ``A.~yes'' and ``B.~no''; score levels keep their order. The prediction is a softmax
over the logits of the valid letters only, at the single answer position; no token is sampled. Questions with more than
26 options are answered as a tournament: a winner per group of 26 options, then a final among the winners. LoRA
\citep{hu2021lora} with rank 16 and $\alpha=32$ is applied to the query, key, value and output projections of the
language model's attention; the 11,489,280 adapter parameters (0.04\% of the model) are trained and everything else
is frozen. The release bundles Gemma's unchanged weights with the adapter, which is applied at load time; merging the
adapter into bf16 weights changed 1.7\% of answers through rounding and was therefore not used. One forward pass is
needed per question.

\subsection{\bevw{}}\label{sec:webmodel}

A browser agent decides at every step which operation to perform and on which element. We express this as two typed
\texttt{choice} questions over a JSON description of the page (URL, title, visible text, numbered interactive elements
with their roles and values, and the recent actions): an \emph{operation} question (click, type text, select, wait,
done, or blocked) and, for each operation that needs one, a \emph{target} question over the eligible elements. This is
the decision interface of the open source browser loop jev-ultrafast \citep{jevultrafast}, into which we plug the model;
the words typed into a field are produced separately by a small text model in the loop. \bevw{} is \bevp{}'s adapter
trained further on generated web steps (Section~\ref{sec:trainweb}); the base model and the document behaviour are
unchanged.

\section{Training data}\label{sec:data}

The models are trained on typed decision cases: public text classification, reading, reasoning and request safety
datasets recast as typed questions about a document, and cases from our own rule based generators of operations
documents (invoices, logs, alerts, tickets) and of written rules applied to documents. Training cases are selected
deterministically and checked against every test set: exact copies and near copies (13-word shingle overlap of 50\% or
more) of any test text are removed, families of near-identical cases are never split between training and evaluation,
and generated test cases use formats, templates and rule wordings that never occur in training. Sources whose terms do
not permit commercial use are excluded from the released models. The training data are not released.

\section{Training}\label{sec:training}

\subsection{\bevl{}}

\paragraph{Objective.} For each question, cross entropy against the label (binary for \texttt{noul}; soft targets where
the source provides a label distribution), computed at each of the three recursive steps and weighted by the halting
probabilities; plus $0.01\times\mathrm{KL}(\text{halting}\,\|\,\text{geometric prior})$; plus $0.5\times$ a selector loss
that pushes the yes/no selector of Eq.~\ref{eq:yesno} towards facts whose exact bit agrees with the label. Within each
case the order of questions is shuffled and a random subset is sometimes kept, so answers cannot depend on position.

\paragraph{Round 1.} From ModernBERT large, one pass over the training mix, in which a few small, high-value sources are
repeated: 18,060 optimiser steps. Batches of 16 cases with 4 accumulation steps (64 cases per step); cases are
shuffled, then sorted by length within windows of 64 batches and the batches shuffled again. AdamW, weight decay 0.01,
learning rate $3\times10^{-5}$ for the encoder and $10^{-3}$ for the decision network, 5\% linear warm up then linear
decay to zero, gradient clipping 1.0, inputs of 4,096 tokens, bf16 autocast with fp32 weights and optimiser state, seed 1.

\paragraph{Round 2.} From round 1: cases not seen in round 1 together with a replay of earlier data, 8,084 steps,
encoder learning rate $2\times10^{-5}$, otherwise as round 1. The released
model is the checkpoint with the best accuracy on a fixed validation mix (step 6,000). After round 1 the model scored
93.7\% on Beveridge Bench but 71.2\% on the typed decisions test; round 2 raised the latter to 75.8\% (and Aegis~2.0 from
80.7\% to 82.3\%) at a cost of 0.7 points on Beveridge-Bench, so more data rather than a different architecture closed
most of the gap.

\subsection{\bevp{}}

200,000 question examples from the same training mix (questions with more than 26 options are not used in training). Loss: cross entropy over the valid option letters at the
answer position. AdamW, learning rate $10^{-4}$, 3\% warm up then linear decay, micro batches of up to 16,384 padded
tokens with 4 accumulation steps (1,560 optimiser steps), inputs up to 4,096 tokens, bf16. The adapter with the best
accuracy on 1,500 validation questions, checked every 200 steps, is kept.

\subsection{\bevw{}}\label{sec:trainweb}

\paragraph{Web step data.} A rule-based generator renders web pages of several site types and the correct next step
towards a written goal. Steps fall into seven kinds: type the query, pick the matching autocomplete suggestion, submit
the search, apply each requested filter, wait while the page loads, open the requested result, and stop when the goal is
met. The test pages come from four site types never used in training (cars, courses, events, libraries).

\paragraph{Recipe.} The released \bevp{} adapter is trained further on generated web steps mixed with document cases
(to keep document skills), learning rate $5\times10^{-5}$, otherwise as \bevp{}. The run stopped at its time limit
after 881 of 992 planned steps; checkpoint selection used the general validation mix, which contains no web steps, and
kept step 200.

\subsection{Calibration}\label{sec:calib}

After training, one temperature per question type is chosen on the validation mix by grid search
($T\in\{0.5,0.7,0.85,1.0,1.2,1.5,2.0,3.0\}$) to minimise negative log likelihood \citep{guo2017calibration}; logits
are divided by $T$ before the softmax. Selected temperatures (yes/no, choice, score): \bevl{} 1.5, 1.2, 1.0; \bevp{}
1.0, 1.2, 1.2; \bevw{} 1.2, 1.0, 1.0. All training ran on a single NVIDIA H200 GPU.

\section{Evaluation setup}\label{sec:eval}

\paragraph{Test sets.} All test sets are held out and were used once, after training. \emph{Decisions on documents},
Beveridge Bench (2,700 generated operations documents, 5,700 questions of 19 kinds, invoices, security logs, service
alerts, customer tickets, counting, dates, amounts, choosing among about 150 options and intrusion in long logs, formats
and templates unseen in training), a public typed decisions test (2,000 questions), Rules Bench (1,000 questions in
which a written rule is applied to a document; rule wordings unseen in training), JailbreakBench refusal decisions \citep{chao2024jailbreakbench}
(200) and Aegis~2.0 request safety \citep{ghosh2025aegis2} (2,538). \emph{Public classification} (500 each): AG News
\citep{zhang2015charcnn}, Emotion \citep{saravia2018carer}, Banking77 \citep{casanueva2020banking77}, SST-5
\citep{socher2013sst}, XNLI English \citep{conneau2018xnli} and MASSIVE intents \citep{fitzgerald2022massive}, none is in
training, although related intent and emotion tasks are. \emph{Areas the models were not built for} (500 each, poems
209), product reviews \citep{zhang2015charcnn}, unfair contract terms and case holdings \citep{chalkidis2022lexglue,
zheng2021casehold}, financial-news sentiment and topics \citep{zeroshot2023finnews}, poem sentiment \citep{sheng2020poem},
support intents \citep{bitext2024support} and toxic comments \citep{borkan2019civil}. None of their texts appears in
training (exact and 13-gram checks).

\paragraph{Baselines.} Jev \citep{typesafejev}, a commercial System One model, through its API with the identical
request. Five general LLMs through Ollama Cloud in October 2026, gpt oss 120B \citep{openai2025gptoss}, Gemma~4 31B
\citep{gemma4_31b}, Mistral Large~3 (675B) and the Mistral Large~4 preview \citep{mistrallarge}, and Nemotron~3 Super
\citep{nemotron3}. Each LLM receives one request per test case containing the document and all its questions, is asked
for a probability for every option in a fixed JSON shape, at temperature 0, with thinking
disabled (gpt-oss: low reasoning effort). Replies are parsed with one shared parser; probabilities are clipped at zero and
renormalised; an unparseable reply is retried once and otherwise scored as wrong.

\paragraph{Metrics.} \emph{Accuracy} is correct answers divided by questions. \emph{Expected calibration error} (ECE) is
$\sum_b \frac{n_b}{N}\,|\mathrm{acc}_b-\mathrm{conf}_b|$ over 10 equal-width confidence bins \citep{naeini2015ece}; for
example a bin holding 40\% of the answers at mean confidence 0.95 of which 90\% are right contributes
$0.4\times0.05=0.02$. \emph{Automation at a target accuracy}, answers are sorted by confidence, most confident first, and
handed over in that order while the accuracy of the handed-over set stays at or above the target, answers with identical
confidence are handed over together, so a system whose fully confident answers are only 94\% right automates 0\% at a
95\% target. It is computed on 10,238 answers per system (Beveridge-Bench, typed decisions, Aegis 2.0).
\emph{Option order flip rate}: the share of answers that change when the options of each question are shuffled.

\section{Results}\label{sec:results}

\subsection{Accuracy}

\begin{table}[htbp]
\centering\small\setlength{\tabcolsep}{3.4pt}
\resizebox{\linewidth}{!}{%
\begin{tabular}{lrrrrrrrr}
\toprule
Test & Large-C & Pro-C & Jev & Mistral L4 & Mistral L3 & gpt-oss 120B & Gemma 4 31B & Nemotron 3 \\
\midrule
\multicolumn{9}{l}{\emph{Decisions on documents}}\\
Beveridge-Bench & \best{93.0} & 91.9 & 87.9 & 83.7 & 74.6 & 91.1 & 88.9 & 70.6 \\
Typed decisions & 75.8 & \best{79.2} & 73.8 & 72.0 & 68.0 & 68.8 & 75.0 & 62.7 \\
Rules-Bench & 85.6 & 92.0 & 92.3 & 83.1 & 81.4 & \best{94.5} & 91.6 & 82.7 \\
JailbreakBench & 84.5 & 84.0 & 83.5 & 77.5 & 62.0 & 68.0 & \best{87.5} & 65.0 \\
Aegis 2.0 & \best{82.3} & 78.2 & 71.5 & 70.5 & 66.9 & 69.5 & 73.1 & 69.3 \\
\midrule
\multicolumn{9}{l}{\emph{Public classification}}\\
AG News & 86.6 & 90.6 & \best{91.8} & 90.8 & 88.8 & 88.8 & \best{91.8} & 87.8 \\
Emotion & 53.2 & 56.2 & 57.6 & 55.8 & 55.6 & \best{57.8} & 55.4 & 50.2 \\
Banking77 & 63.6 & 70.6 & 79.2 & 75.8 & 71.2 & \best{81.0} & 78.8 & 69.2 \\
SST-5 & 33.2 & 51.8 & \best{56.8} & 51.4 & 49.4 & 49.6 & 53.2 & 46.8 \\
XNLI (en) & 84.6 & 86.4 & \best{89.2} & 84.8 & 77.6 & 80.2 & \best{89.2} & 75.0 \\
MASSIVE & 59.4 & 74.8 & 80.8 & 78.4 & 75.2 & 81.6 & \best{83.8} & 69.4 \\
\midrule
\multicolumn{9}{l}{\emph{Areas not built for}}\\
Product reviews & 93.2 & 96.4 & 96.4 & 96.4 & 96.6 & 95.4 & \best{97.2} & 94.8 \\
Unfair contract terms & 65.6 & \best{76.0} & 62.0 & 54.6 & 52.8 & 47.6 & 74.4 & 44.0 \\
Financial sentiment & 69.8 & 59.2 & 66.4 & 64.0 & 50.4 & 68.4 & \best{71.2} & 64.0 \\
Poem sentiment & \best{71.8} & 65.1 & 44.5 & 42.6 & 32.1 & 59.8 & 62.7 & 45.5 \\
Support intents & 85.6 & 88.0 & 89.8 & 85.4 & 86.2 & 90.0 & \best{90.4} & 76.6 \\
Toxic comments & 63.0 & 79.2 & 78.8 & 68.0 & 73.0 & \best{80.8} & 76.4 & 69.6 \\
Financial topics & 37.8 & \best{69.8} & 65.0 & 61.8 & 61.6 & 63.4 & 64.2 & 57.2 \\
Case holdings & 42.0 & 71.2 & \best{77.6} & 69.2 & 72.2 & 61.2 & 71.6 & 55.0 \\
\bottomrule
\end{tabular}}
\caption{Accuracy (\%) on held out tests, identical cases for every system. Bold: best in the row.}
\label{tab:main}
\end{table}

\begin{figure}[htbp]
\centering
\includegraphics[width=\linewidth]{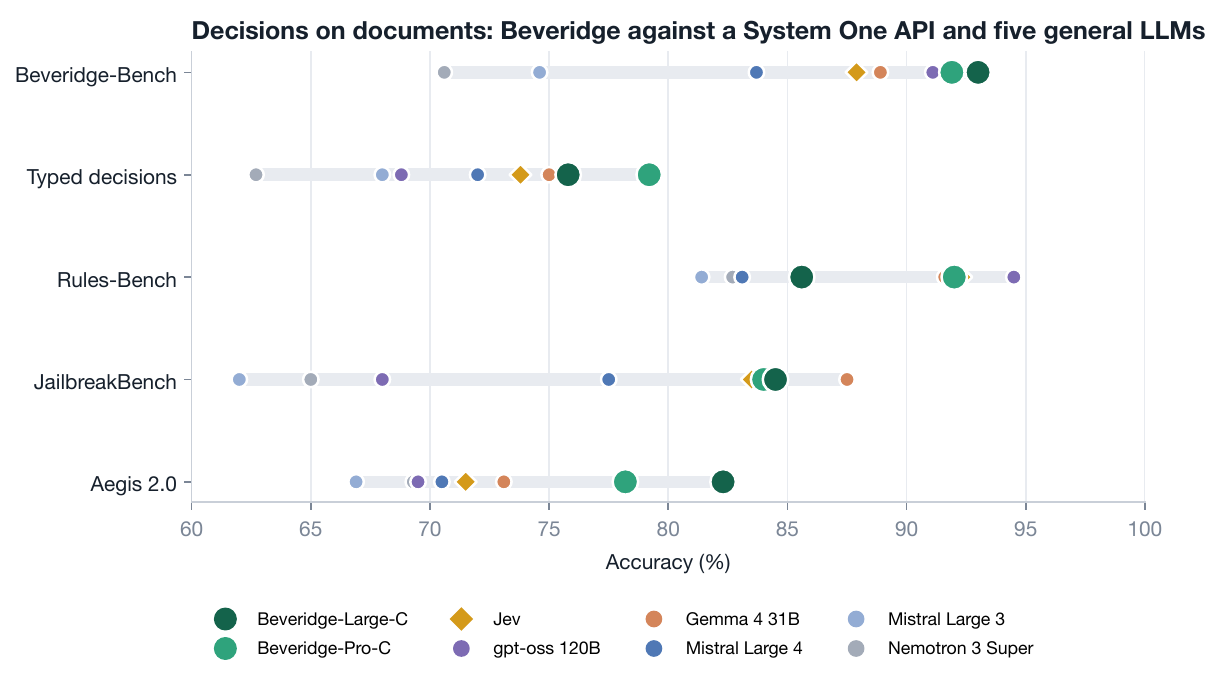}
\caption{Accuracy on the five document decision tests. Large markers show the Beveridge models, the diamond shows Jev
and small markers show the general LLMs.}
\label{fig:overview}
\end{figure}

\begin{figure}[htbp]
\centering
\includegraphics[width=0.9\linewidth]{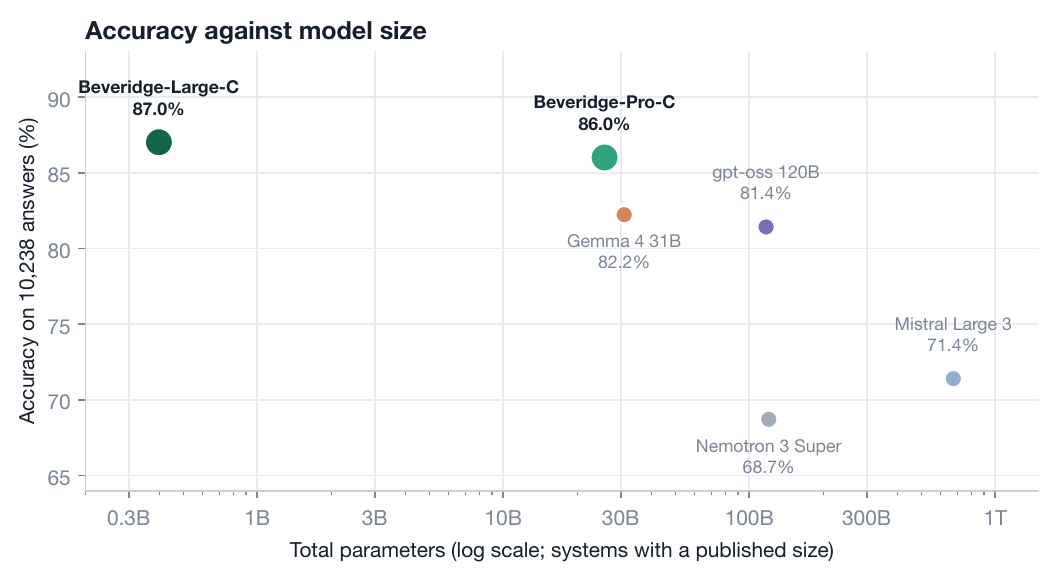}
\caption{Accuracy on 10,238 held out answers (Beveridge Bench, typed decisions, Aegis~2.0) against total parameters,
for every system with a published size.}
\label{fig:size}
\end{figure}

Table~\ref{tab:main} gives accuracy on all 19 comparison tests. On decisions about operations documents, the 0.4B
\bevl{} is the most accurate system on Beveridge-Bench (93.0\%) and Aegis~2.0 (82.3\%), ahead of every general LLM
including models up to 675B parameters. \bevp{} is best on the typed-decisions test (79.2\%) and, among Beveridge models,
on rules and knowledge-heavy tasks: it gains 32.0 points over \bevl{} on financial news topics and 29.2 on legal case
holdings, which require knowledge rather than reading. General LLMs and Jev remain stronger on large intent sets
(Banking77, MASSIVE), five-level sentiment and case holdings. Figure~\ref{fig:overview} shows every system on the
document tests and Figure~\ref{fig:size} places accuracy against model size.

\subsection{Per question behavior}

\begin{figure}[htbp]
\centering
\includegraphics[width=0.78\linewidth]{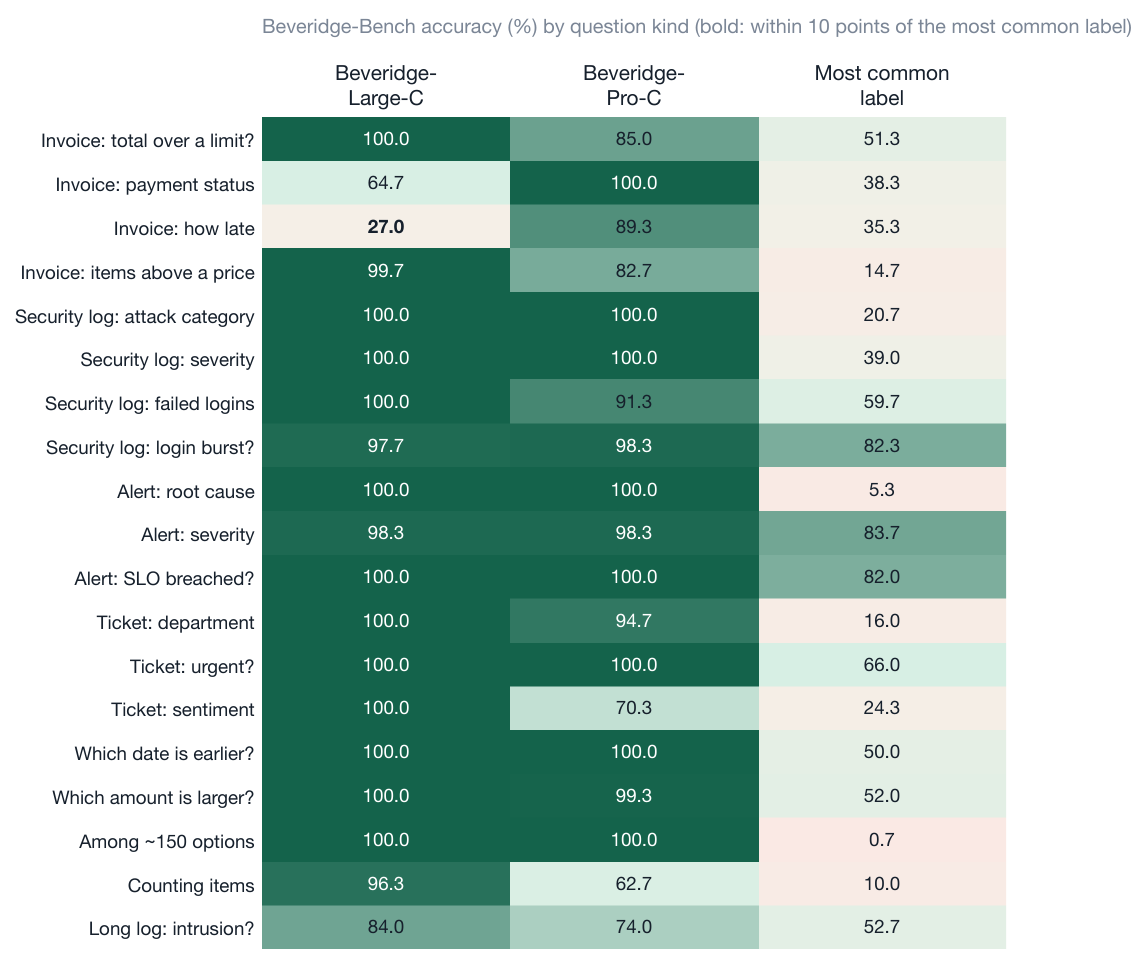}
\caption{Beveridge Bench accuracy (\%) by question kind, 300 cases each, with the accuracy of always answering the most
common label (the data of Table~\ref{tab:perq} as a heatmap).}
\label{fig:perq}
\end{figure}

Averages hide weak spots. Table~\ref{tab:perq} and Figure~\ref{fig:perq} break Beveridge Bench into its 19 question kinds with the accuracy of
always answering the most common label. \bevl{} scores 96\% or more on 16 of the 19 kinds but fails on \emph{how late an invoice
was paid}: 27.0\%, below the 35.3\% majority baseline, and is weak on payment status against the due date (64.7\%); both
require date arithmetic against ``today'', which the rule engine does not compute. \bevp{} solves both (89.3\% and
100\%) but is weaker on counting (62.7\%) and ticket sentiment (70.3\%). The models are complementary, which motivates
routing (Section~\ref{sec:discussion}).

\begin{table}[htbp]
\centering\small
\begin{tabular}{lrrr}
\toprule
Question (300 cases each) & Large-C & Pro-C & Majority \\
\midrule
Invoice: total over a limit? & 100.0 & 85.0 & 51.3 \\
Invoice: payment status vs due date & 64.7 & 100.0 & 38.3 \\
Invoice: how late & 27.0 & 89.3 & 35.3 \\
Invoice: items above a price & 99.7 & 82.7 & 14.7 \\
Security log: attack category & 100.0 & 100.0 & 20.7 \\
Security log: severity & 100.0 & 100.0 & 39.0 \\
Security log: failed-login count & 100.0 & 91.3 & 59.7 \\
Security log: login burst? & 97.7 & 98.3 & 82.3 \\
Alert: root cause & 100.0 & 100.0 & 5.3 \\
Alert: severity & 98.3 & 98.3 & 83.7 \\
Alert: SLO breached? & 100.0 & 100.0 & 82.0 \\
Ticket: department & 100.0 & 94.7 & 16.0 \\
Ticket: urgent? & 100.0 & 100.0 & 66.0 \\
Ticket: sentiment & 100.0 & 70.3 & 24.3 \\
Which date is earlier? & 100.0 & 100.0 & 50.0 \\
Which amount is larger? & 100.0 & 99.3 & 52.0 \\
Choosing among about 150 options & 100.0 & 100.0 & 0.7 \\
Counting items in a list & 96.3 & 62.7 & 10.0 \\
Long log: intrusion? & 84.0 & 74.0 & 52.7 \\
\bottomrule
\end{tabular}
\caption{Beveridge Bench accuracy (\%) by question kind.}
\label{tab:perq}
\end{table}

\subsection{Ablations}\label{sec:ablation}

\begin{figure}[htbp]
\centering
\includegraphics[width=0.9\linewidth]{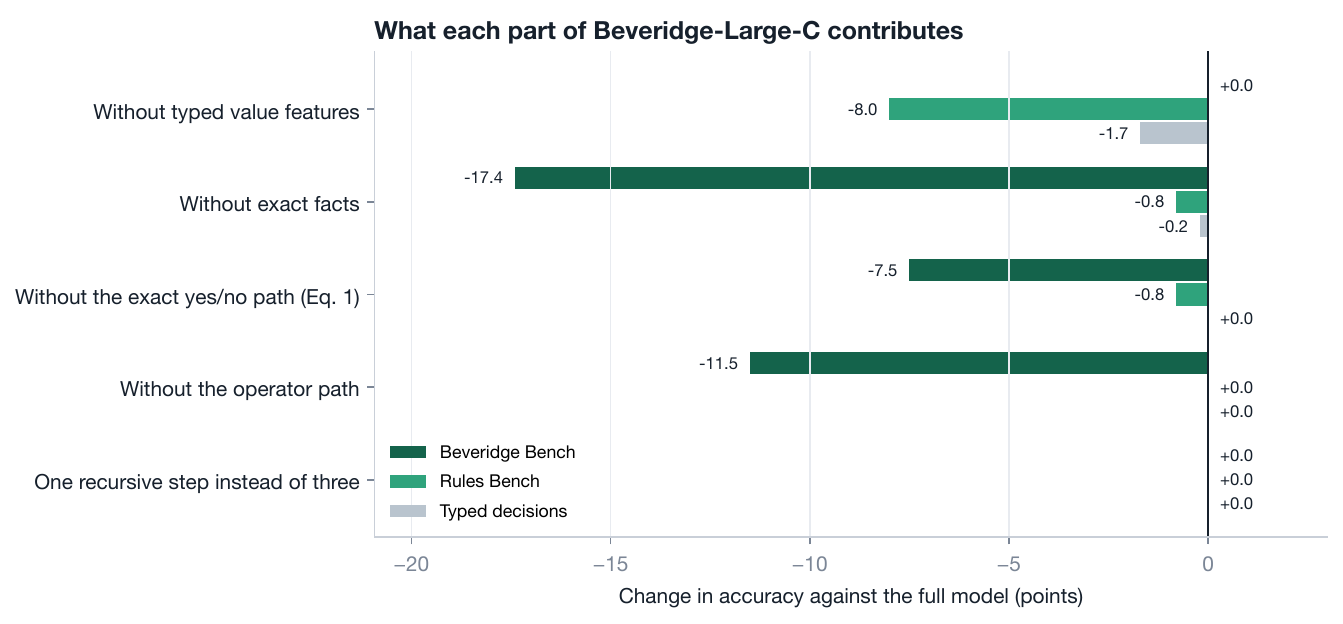}
\caption{Change in accuracy when one part of \bevl{} is switched off at inference time, against the full model, on Beveridge Bench, Rules Bench and typed decisions.}
\label{fig:ablation}
\end{figure}

To measure how much the released model relies on each part, we switch one part off at inference time and score the same held out tests again (Table~\ref{tab:ablation}, Figure~\ref{fig:ablation}). The weights are unchanged, so the numbers show how much the trained model relies on a part, not what a model trained without it would reach.

The exact facts carry Beveridge Bench, without them accuracy falls from 93.0\% to 75.6\% (17.4 points) and ECE rises from 0.029 to 0.067. Two paths that use the facts explain much of this, switching off the operator path (lexical option features and the match between an option's number and a computed count) costs 11.5 points and switching off the exact yes/no path (Eq.~\ref{eq:yesno}) costs 7.5 points (the two overlap). The typed value features matter most when a written rule compares values, without them Rules Bench falls from 85.6\% to 77.6\% and its ECE rises from 0.034 to 0.075, while Beveridge Bench is unchanged. On the typed decisions test no part changes accuracy by more than 1.7 points.

The recursive core is the exception, with one step instead of three, accuracy is unchanged on all three tests (to one decimal) and ECE moves by at most 0.001. In the released model the later steps therefore add nothing measurable to the answers of the first, and the three step core works in practice as a single transformer layer over questions and facts. We report the recursion as part of the architecture, not as a source of accuracy.

Further variants (using only the last of the three steps and removing temperature calibration) were still running when this version was prepared and will be added in an update.

\begin{table}[htbp]
\centering\small
\resizebox{\linewidth}{!}{%
\begin{tabular}{lrrrrrr}
\toprule
 & \multicolumn{2}{c}{Beveridge Bench} & \multicolumn{2}{c}{Rules Bench} & \multicolumn{2}{c}{Typed decisions}\\
\cmidrule(lr){2-3}\cmidrule(lr){4-5}\cmidrule(lr){6-7}
Variant & Acc. (change) & ECE & Acc. (change) & ECE & Acc. (change) & ECE\\
\midrule
Full model (released) & 93.0 & 0.029 & 85.6 & 0.034 & 75.8 & 0.170\\
\midrule
Without typed value features & 93.0 (+0.0) & 0.027 & 77.6 (-8.0) & 0.075 & 74.2 (-1.7) & 0.152\\
Without exact facts & 75.6 (-17.4) & 0.067 & 84.8 (-0.8) & 0.041 & 75.6 (-0.2) & 0.171\\
Without the exact yes/no path (Eq.~\ref{eq:yesno}) & 85.6 (-7.5) & 0.041 & 84.8 (-0.8) & 0.043 & 75.8 (+0.0) & 0.170\\
Without the operator path & 81.5 (-11.5) & 0.025 & 85.6 (+0.0) & 0.034 & 75.8 (+0.0) & 0.170\\
One recursive step instead of three & 93.0 (+0.0) & 0.030 & 85.6 (+0.0) & 0.034 & 75.8 (+0.0) & 0.169\\
\bottomrule
\end{tabular}}
\caption{Inference time ablations of \bevl{}: accuracy (\%) with the change from the full model in points, and ECE. 5,700, 1,000 and 2,000 questions. The trained weights are unchanged; each row switches one part off.}
\label{tab:ablation}
\end{table}

\subsection{Calibration and automation}

\begin{table}[htbp]
\centering\small
\begin{tabular}{lrrrr|rrrr}
\toprule
& \multicolumn{4}{c|}{\bevl{}} & \multicolumn{4}{c}{\bevp{}} \\
Test & Acc. & Cov.@0.8 & Acc.@0.8 & ECE & Acc. & Cov.@0.8 & Acc.@0.8 & ECE \\
\midrule
Beveridge-Bench & 93.0 & 86.9 & 98.9 & 0.029 & 91.9 & 89.7 & 95.0 & 0.032 \\
Rules-Bench & 85.6 & 76.8 & 94.7 & 0.034 & 92.0 & 87.2 & 97.0 & 0.011 \\
Aegis 2.0 & 82.3 & 57.0 & 94.2 & 0.034 & 78.2 & 51.1 & 92.6 & 0.024 \\
Product reviews & 93.2 & 84.2 & 96.9 & 0.024 & 96.4 & 93.0 & 97.9 & 0.014 \\
Support intents & 85.6 & 72.6 & 95.0 & 0.031 & 88.0 & 98.8 & 88.3 & 0.105 \\
Typed decisions & 75.8 & 14.0 & 97.1 & 0.170 & 79.2 & 61.7 & 91.5 & 0.040 \\
\bottomrule
\end{tabular}
\caption{Calibration. Cov.@0.8, share of answers given with confidence $\ge 0.8$, Acc.@0.8: their accuracy.}
\label{tab:calib}
\end{table}

\begin{table}[htbp]
\centering\small
\begin{tabular}{lrrr}
\toprule
System & Automated at 95\% & Automated at 90\% & Accuracy, all answers \\
\midrule
\bevl{} & \best{76\%} & \best{93\%} & \best{87.0} \\
\bevp{} & 69\% & 90\% & 86.0 \\
Jev & 59\% & 77\% & 81.1 \\
gpt-oss 120B & 0\% & 74\% & 81.4 \\
Gemma 4 31B & 0\% & 71\% & 82.2 \\
Mistral Large 4 & 0\% & 0\% & 78.1 \\
Mistral Large 3 & 0\% & 0\% & 71.4 \\
Nemotron 3 Super & 0\% & 0\% & 68.7 \\
\bottomrule
\end{tabular}
\caption{Share of decisions that can be handed over while the handed-over decisions stay at least 95\% (90\%) correct,
on 10,238 answers per system.}
\label{tab:auto}
\end{table}

\begin{figure}[htbp]
\centering
\includegraphics[width=0.95\linewidth]{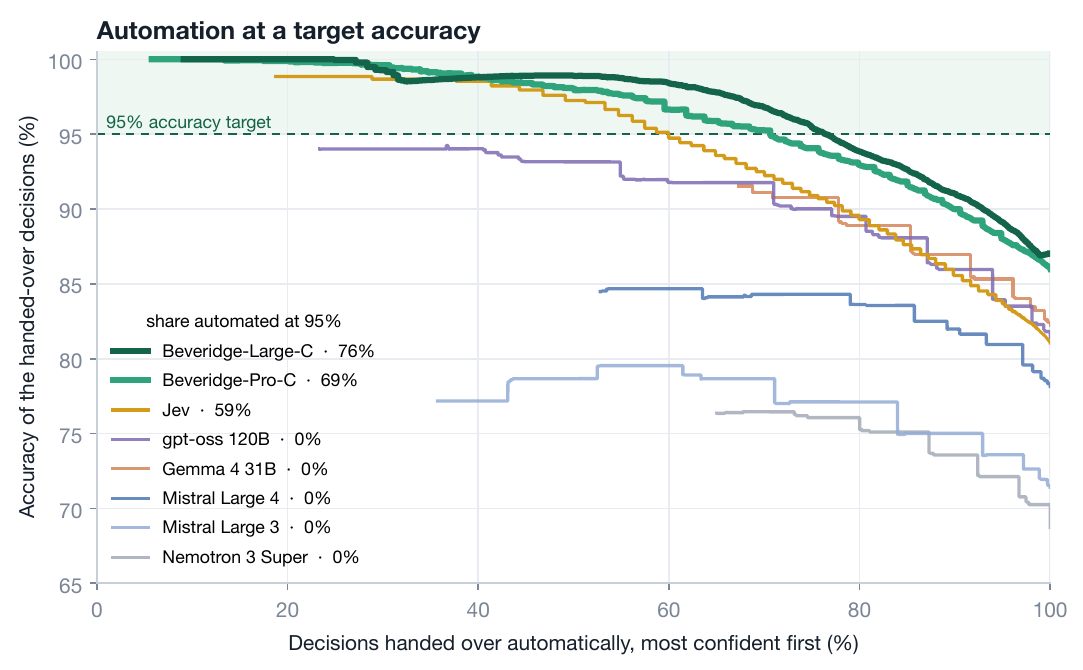}
\caption{Accuracy of the automated decisions against the share handed over, most confident first.}
\label{fig:auto}
\end{figure}

\begin{figure}[htbp]
\centering
\includegraphics[width=0.95\linewidth]{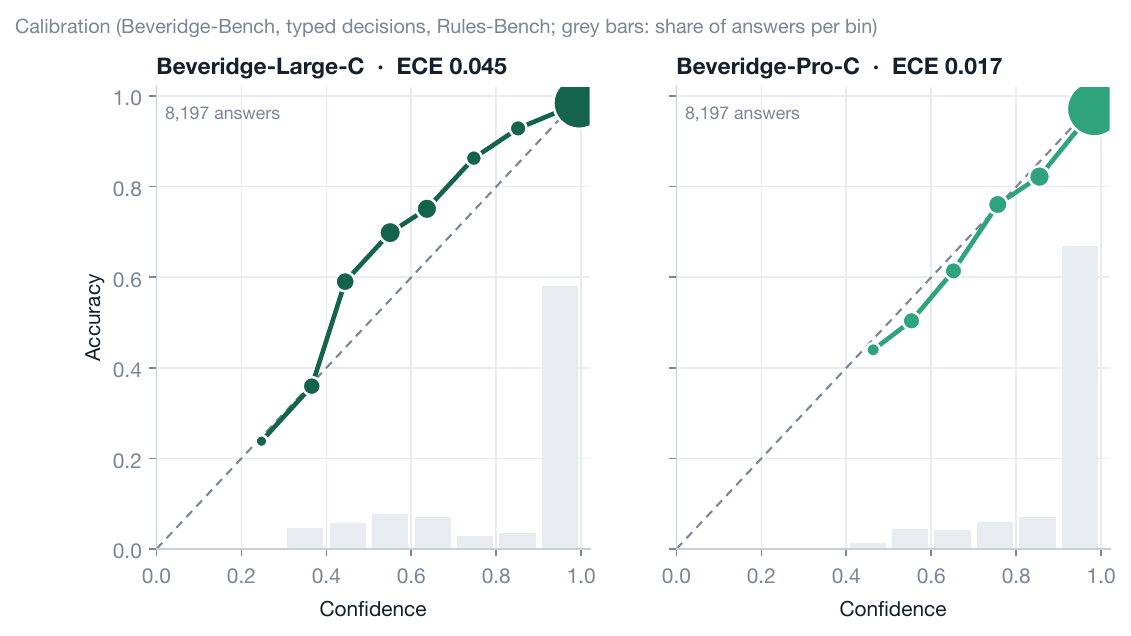}
\caption{Reliability diagrams pooled over Beveridge Bench, typed decisions and Rules Bench (questions with up to 50
options, 8,197 answers per model). Dot size shows the share of answers in the bin, the dashed line is perfect
calibration.}
\label{fig:rel}
\end{figure}

On most tests the Beveridge models' confidences track their accuracy (Table~\ref{tab:calib}, ECE 0.011--0.034,
pooled reliability diagrams in Figure~\ref{fig:rel}).

 The two
exceptions are informative, \bevl{} is \emph{under}-confident on the typed-decisions test (ECE 0.170, only 14.0\% of
answers above 0.8 but 97.1\% of those correct), and \bevp{} is \emph{over}-confident on support intents (98.8\% of
answers above 0.8 at 88.3\% accuracy), a domain outside its training.

Calibration is what turns accuracy into automation (Table~\ref{tab:auto}, Figure~\ref{fig:auto}). The general LLMs give
the same, maximal confidence to a large block of their answers (between 23\% and 67\% of them, Figure~\ref{fig:auto});
because answers with equal confidence must be handed over together, and each of these blocks is less than 95\% correct,
none of the five LLMs automates any decision at a 95\% target. \bevl{} automates 76\% of decisions at 95\% and 93\% at 90\%; Jev, the other System One
model, automates 59\%. The option order flip rate of \bevl{} is 3.4\% (103 of 3,035 answers change when options are
shuffled).

\subsection{Speed}

\begin{figure}[htbp]
\centering
\includegraphics[width=0.9\linewidth]{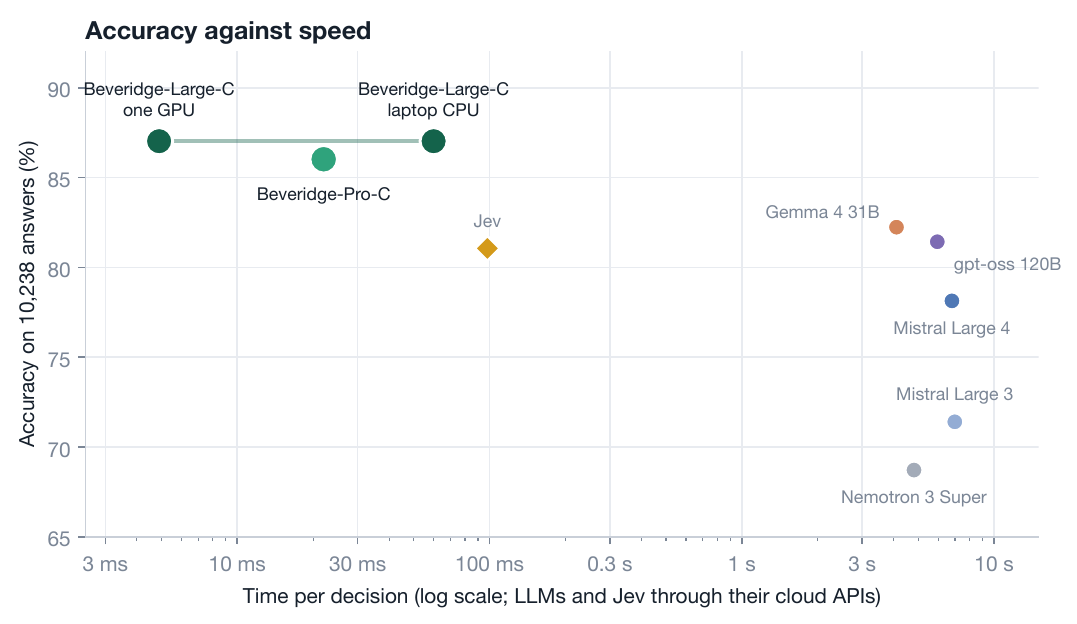}
\caption{Accuracy on 10,238 held out answers against time per decision. The Beveridge models were measured with the
released package on the hardware shown, Jev and the LLMs by median time through their cloud APIs, including network and
queueing.}
\label{fig:speed}
\end{figure}

Table~\ref{tab:speed} reports time per decision with the released package on a document with three questions.
\bevl{} answers in 4.9\,ms on one H200 GPU (3.4\,ms in batches of 64), 17\,ms on the 20 core GPU of an Apple M4 Pro laptop and
60\,ms on its 14 core CPU (10 performance and 4 efficiency cores, 24\,GB memory, PyTorch's default of 10 threads). \bevp{} needs 22\,ms (7.8\,ms batched) on one H200. For reference, the median time per decision through the
cloud APIs, including network and queueing, was 98\,ms for Jev and 4.1--7.0\,s for the five LLMs (Figure~\ref{fig:speed}).

\begin{table}[htbp]
\centering\small
\begin{tabular}{llrr}
\toprule
Model & Hardware & Per decision & Batched (64) \\
\midrule
\bevl{} & NVIDIA H200 & 4.9 ms & 3.4 ms \\
\bevl{} & Apple M4 Pro laptop GPU (20 cores) & 17 ms & \\
\bevl{} & Apple M4 Pro laptop CPU (14 cores) & 60 ms & \\
\bevp{} & NVIDIA H200 & 22 ms & 7.8 ms \\
\bottomrule
\end{tabular}
\caption{Time per decision, one document with three questions, released package.}
\label{tab:speed}
\end{table}

\subsection{Web navigation}\label{sec:web}

\begin{table}[htbp]
\centering\small
\resizebox{\linewidth}{!}{%
\begin{tabular}{lrr}
\toprule
Test & \bevp{} & \bevw{} \\
\midrule
Next step, 500 pages of 4 unseen site types & 49.4 & \best{98.2} \\
Next operation, same pages & 60.2 & \best{98.4} \\
24 recorded real browser steps: same operation as the reference agent & 22/24 & \best{24/24} \\
Live hotel search, local test page (3 runs) & 0/3 & \best{3/3}, 1.86 s \\
Live Wikipedia article search (3 runs) & 3/3, 4.9 s & \best{3/3}, 4.88 s \\
Beveridge-Bench (documents) & 91.9 & \best{92.1} \\
\bottomrule
\end{tabular}}
\caption{Web navigation (accuracy \%, or successful runs with median time from the first decision to completion, one
H200). The reference agent is Jev in jev-ultrafast; its published live times are 1.2\,s and 3.2\,s on different hardware.}
\label{tab:web}
\end{table}

\begin{figure}[htbp]
\centering
\includegraphics[width=0.9\linewidth]{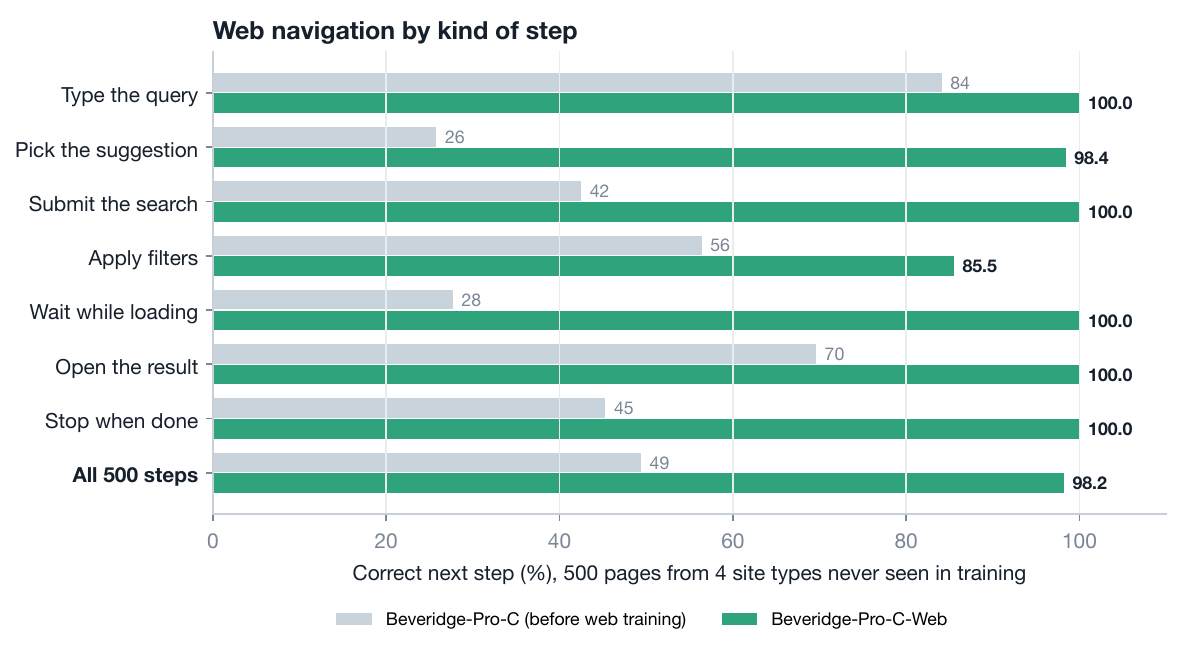}
\caption{Correct next step by kind of step on 500 pages from four site types never seen in training, before and after
web training.}
\label{fig:web}
\end{figure}

Before web training, \bevp{} already agreed with the reference agent on 22 of 24 recorded real browser steps but chose
the correct next step on only 49.4\% of generated pages from unseen site types; its main failures were picking
autocomplete suggestions, waiting for loading pages and pressing search before applying filters, and in live use it
applied filters before submitting the search. \bevw{} reaches 98.2\% on the held out pages (by kind of step: typing,
submitting, waiting, opening the result and stopping 100\%, autocomplete 98.4\%, filters 85.5\%), agrees with the
reference agent on all 24 real steps, and completes both live tasks in all runs (Table~\ref{tab:web}, Figure~\ref{fig:web}), with about
0.2\,s per decision on the hotel page and 0.7\,s on longer Wikipedia pages. Its Beveridge-Bench accuracy is unchanged
(92.1\% against 91.9\%), so one model serves documents and browsing. Two live tasks are a small test, and both
resemble site types represented in training (travel stays, encyclopedia search); we treat them as a demonstration, and the
500 pages of unseen site types as the measure of generalisation.

\subsection{Reproducibility of the release}

The released package reproduces the evaluation run exactly for \bevl{} (3,000 of 3,000 answers identical). Loading each
released model as a user would (fresh download from Hugging Face, released package) gives 93.04\% on Beveridge Bench
for \bevl{} (identical to the evaluation), 92.04\% for \bevp{} (evaluation 91.89\%; a few near-tied answers of the
mixture-of-experts model change with batch composition in bf16) and 92.09\% for \bevw{}.

\section{Discussion}\label{sec:discussion}

\paragraph{When a small encoder wins.} Decisions that hinge on reading values in a document and comparing them
(limits, counts, dates, categories defined by written criteria) are where \bevl{} leads. Two design choices carry this:
the model sees numbers as numbers (value features) and can defer to exact computations (Eq.~\ref{eq:yesno}). Where the
answer needs world knowledge (legal holdings, financial topics) or very fine sentiment, a large pretrained LM is better,
and \bevp{} recovers most of that gap within the same interface.

\begin{figure}[htbp]
\centering
\begin{tabular}{@{}ccc@{}}
\href{https://bev.sjothi.com/research/media/beveridge-web-browsing.mp4}{\includegraphics[width=0.32\linewidth]{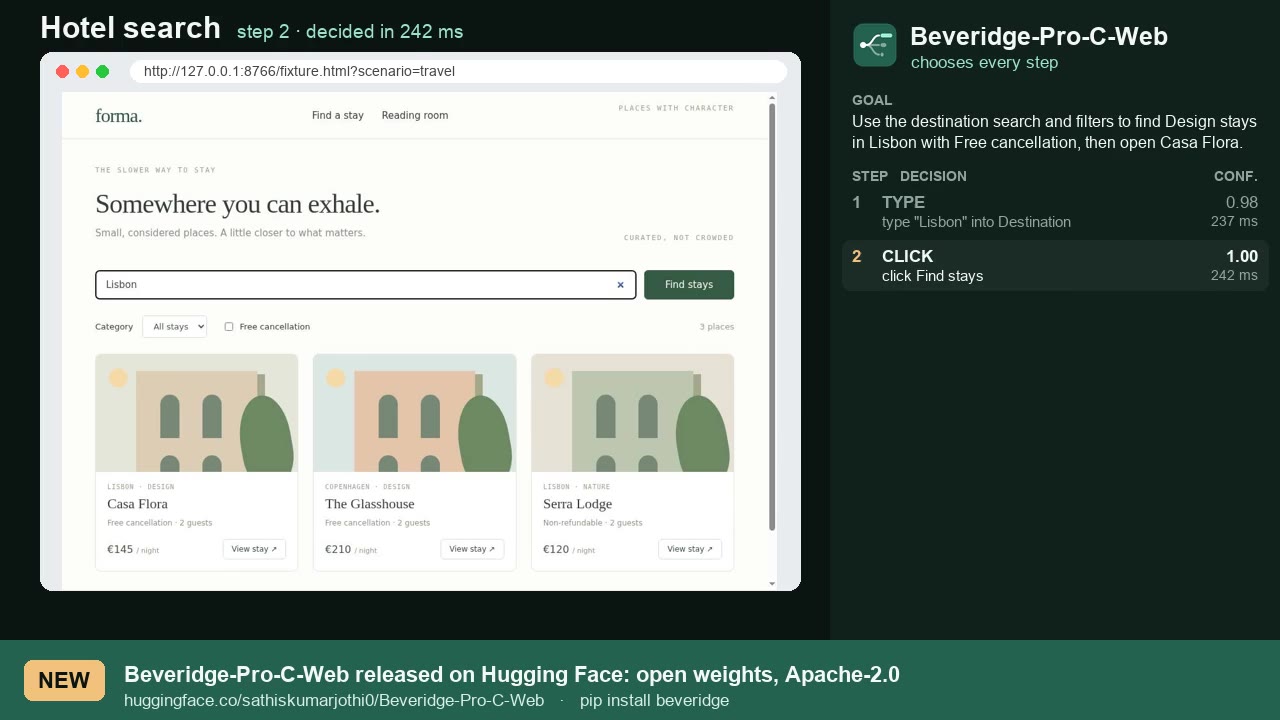}} &
\href{https://bev.sjothi.com/research/media/beveridge-voice-examples.mp4}{\includegraphics[width=0.32\linewidth]{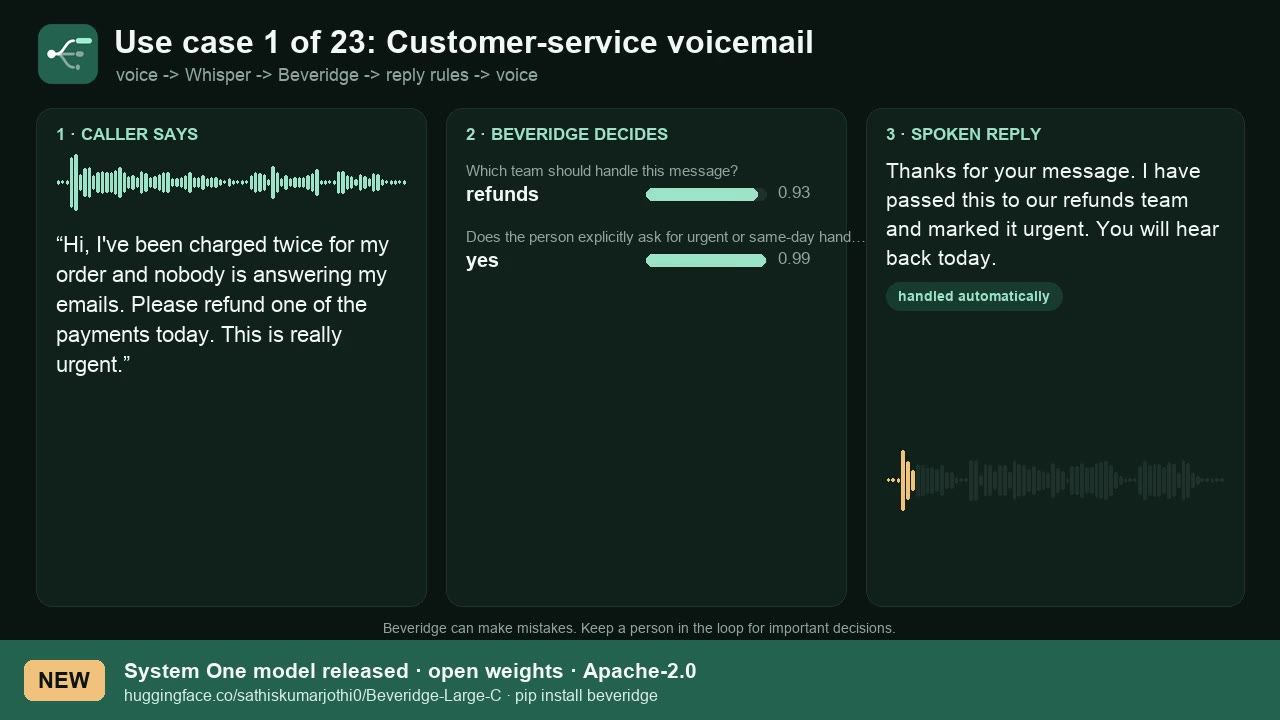}} &
\href{https://bev.sjothi.com/research/media/bev-snake.mp4}{\includegraphics[width=0.32\linewidth]{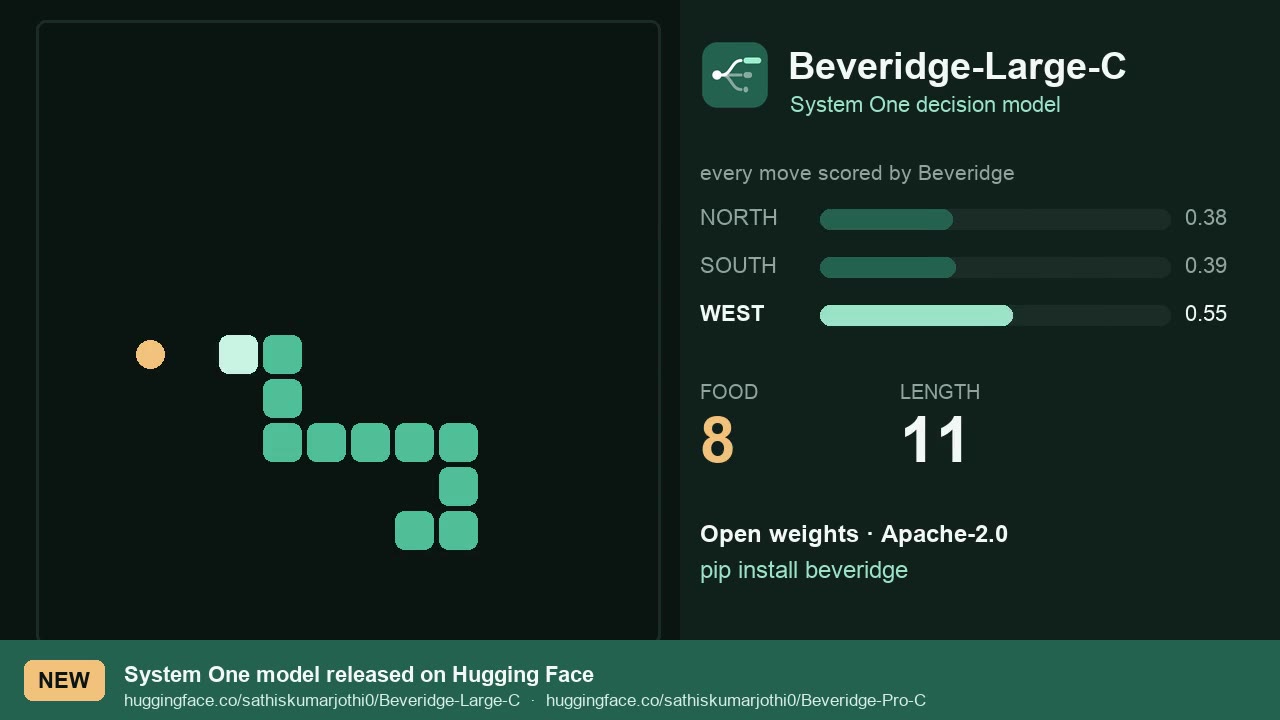}}\\
\small (a) Browsing live & \small (b) Voice assistant & \small (c) Playing Snake
\end{tabular}
\caption{Demonstration videos (click a still or a link to open the video).
(a) \href{https://bev.sjothi.com/research/media/beveridge-web-browsing.mp4}{Browsing live}
(\url{https://bev.sjothi.com/research/media/beveridge-web-browsing.mp4}), \bevw{} choosing every step of a browser agent
(\href{https://huggingface.co/sathiskumarjothi0/Beveridge-Pro-C-Web}{model card}).
(b) \href{https://bev.sjothi.com/research/media/beveridge-voice-examples.mp4}{Voice assistant}
(\url{https://bev.sjothi.com/research/media/beveridge-voice-examples.mp4}), speech recognition transcribes a voice note,
\bevl{} answers the use case's typed questions and fixed reply rules produce the spoken reply (stock synthetic voices).
(c) \href{https://bev.sjothi.com/research/media/bev-snake.mp4}{Playing Snake}
(\url{https://bev.sjothi.com/research/media/bev-snake.mp4}), \bevl{} scoring every move of the game Snake with narrow
typed questions.}
\label{fig:demos}
\end{figure}

\paragraph{Narrow questions.} In applications we found that the models are most reliable when a decision is asked as
narrow, descriptive questions whose answers code then combines (``Which team handles this?'' and ``Does it ask for
same day handling?'') rather than as one vague question (``Is this urgent?''). Our browser, voice assistant and Snake
demonstrations all follow this pattern (Figure~\ref{fig:demos}).

\paragraph{Routing.} The two models are complementary (Table~\ref{tab:perq}) and both are calibrated, so a request can be
answered by \bevl{} first and only questions below a confidence threshold sent to \bevp{}; the released package
implements this as a router.

\paragraph{Why not just prompt an LLM?} The LLM baselines reach similar or higher accuracy on several public tests, but
their stated probabilities are not usable as thresholds (Table~\ref{tab:auto}), each decision costs seconds, and outputs
must be parsed. A System One model gives up open-ended reasoning in exchange for a fixed output space, calibrated
probabilities and millisecond latency.

\section{Limitations}

\begin{itemize}
  \item \textbf{Knowledge heavy decisions}: \bevl{} scores 42.0\% on legal case holdings and 37.8\% on financial news
  topics.
  \item \textbf{Date arithmetic against today}: how late an invoice is paid is not reliable for \bevl{} (27.0\%); compute
  due dates and lateness in code.
  \item \textbf{Large option sets}: with hundreds of options the top answer is often right but its probability is low;
  \bevp{} answers more than 26 options as a tournament with lower accuracy (Banking77 70.6\%).
  \item \textbf{Option order}: 3.4\% of \bevl{}'s answers change when options are shuffled; keep the order stable.
  \item \textbf{Calibration outside the training domains} can be poor (support intents for \bevp{}).
  \item \textbf{Scope}: English only; inputs beyond 4,096 tokens are truncated; no images. Training and evaluation used
  one to five questions per request, accuracy with six to 15 questions per request is untested. Web results cover two live
  tasks and generated pages.
  \item \textbf{Evaluation}: several tests are our own generated suites, general LLMs were evaluated zero-shot through
  one hosting provider with a single prompt format, and their numbers may change with prompting.
  \item \textbf{Use}: the models are not intended for medical, legal or financial advice, for consequential decisions,
  validate on your own data and keep a person in the loop.
\end{itemize}

\section{Release}

Weights are released under Apache-2.0 on Hugging Face:
\href{https://huggingface.co/sathiskumarjothi0/Beveridge-Large-C}{\bevl{}}, \href{https://huggingface.co/sathiskumarjothi0/Beveridge-Pro-C}{\bevp{}} and
\href{https://huggingface.co/sathiskumarjothi0/Beveridge-Pro-C-Web}{\bevw{}}, \bevp{} and \bevw{} contain or load Gemma~4 under its
Apache-2.0 licence, and \bevl{} contains ModernBERT-large (Apache-2.0). Inference code is the \texttt{bev-decisions}
Python package on PyPI (\url{https://pypi.org/project/bev-decisions/}, also installable as \texttt{beveridge},
\url{https://pypi.org/project/beveridge/}), with source and examples on GitHub
(\url{https://github.com/sathiskumarjothi0-oss/bev-decisions-pypi}, version 0.2.5 was used for this paper), with a command-line tool, an HTTP server, an MCP server for AI
assistants and a router. The training datasets are not released.

\section{Conclusion}

Beveridge shows that a 0.4B encoder with typed value features, exact facts and a small recursive decision network can
match or exceed general LLMs up to three orders of magnitude larger on document decisions, while giving calibrated
probabilities that make automation at a target accuracy possible, at millisecond latency on commodity hardware. The same
interface carries over to a LoRA adapted mixture-of-experts LLM for knowledge-heavy decisions and to the next step of a
browser agent.

\paragraph{About the name.} The models are named after Sir William Beveridge (1879--1963), whose 1942 report laid the
foundations of the modern welfare state and the UK National Health Service; the name is a reminder of the goal of
dependable, fair decisions that serve people.

\FloatBarrier

\appendix
\section{Example request and response}\label{app:example}

\begin{quote}\small
\begin{verbatim}
state: "Subject: Charged twice for order 88412. I was billed
        $49.99 twice on 3 October. Please refund one payment today."
questions:
  team:
    type: choice
    instructions: "Which team should handle this ticket?"
    criteria: {billing: "payments, invoices, refunds",
               tech: "bugs, outages, errors",
               sales: "pricing and plans"}
  urgent:
    type: noul
    instructions: "Does the customer ask for urgent handling?"
  severity:
    type: score
    instructions: "How severe is the problem for the customer?"
    criteria: [low, medium, high]
answer:
  team:     {choice, confidence, probabilities over the 3 keys}
  urgent:   {noul: P(yes)}
  severity: {label, score (expected level), probabilities}
\end{verbatim}
\end{quote}

\end{document}